\pdfoutput=1
\documentclass[11pt]{article}

\usepackage[final]{acl}

\usepackage{times}
\usepackage{latexsym}

\usepackage[T1]{fontenc}

\usepackage[utf8]{inputenc}

\usepackage{microtype}

\usepackage{inconsolata}

\usepackage{tcolorbox}
\usepackage{xcolor}
\usepackage{tabularx}
\tcbuselibrary{skins}
\tcbuselibrary{breakable}

\usepackage{multirow}
\usepackage{float}
\usepackage{mathtools}
\usepackage{ifthen}
\usepackage{bm}
\usepackage{fp}
\usepackage{siunitx}
\usepackage{amsthm}
\usepackage{caption}
\usepackage{subcaption}
\usepackage{xspace}
\usepackage{amsmath, amsthm, amssymb}
\usepackage{pifont}
\usepackage{wrapfig}
\usepackage{pifont}
\usepackage{algorithm}
\usepackage{algorithmicx}
\usepackage[noend]{algpseudocode}
\usepackage{comment}
\usepackage{listings}
\usepackage{booktabs}
\usepackage{makecell}

\title{Learning Like Humans: Advancing LLM Reasoning Capabilities via Adaptive Difficulty Curriculum Learning and Expert-Guided Self-Reformulation}

\author{
 \textbf{Enci Zhang\textsuperscript{1,2,3}},
 \textbf{Xingang Yan\textsuperscript{1,2}},
 \textbf{Wei Lin\textsuperscript{1,2}\thanks{Wei Lin is the corresponding author.}},
 \textbf{Tianxiang Zhang\textsuperscript{1,2}},
 \textbf{Qianchun Lu\textsuperscript{1,2}}
\\
\\
 \textsuperscript{1}State Key Laboratory of Mobile Network and Mobile Multimedia Technology, Shenzhen, China\\
 \textsuperscript{2}ZTE Corporation, Shenzhen, China\\
 \textsuperscript{3}Peking University, Shenzhen, China
\\
 \small{
   \textbf{E-mail:} \href{mailto:email@domain}{eczhang@stu.pku.edu.cn}, \href{mailto:email@domain}{weilincs@pku.org.cn}
 }
}

\begin{document}
\maketitle
\begin{abstract}
Despite impressive progress in areas like mathematical reasoning, large language models still face  challenges in consistently solving complex problems. Drawing inspiration from key human learning strategies, we propose two novel strategies to enhance the capability of large language models to solve these complex problems. 
First, Adaptive Difficulty Curriculum Learning (ADCL) is a novel curriculum learning strategy that tackles the \emph{Difficulty Shift} phenomenon (i.e., a model's perception of problem difficulty dynamically changes during training) by periodically re-estimating difficulty within upcoming data batches to maintain alignment with the model's evolving capabilities. 
Second, Expert-Guided Self-Reformulation (EGSR) is a novel reinforcement learning strategy that bridges the gap between imitation learning and pure exploration by guiding models to reformulate expert solutions within their own conceptual framework, rather than relying on direct imitation, fostering deeper understanding and knowledge assimilation. 
Extensive experiments on challenging mathematical reasoning benchmarks, using Qwen2.5-7B as the base model, demonstrate that these human-inspired strategies synergistically and significantly enhance performance. Notably, their combined application improves performance over the standard Zero-RL baseline by 10\% on the AIME24 benchmark and 16.6\% on AIME25.
\end{abstract}

\section{Introduction}
The landscape of complex reasoning in large language models (LLMs) has been dramatically reshaped by recent breakthroughs, exemplified by models such as OpenAI-o1\cite{intro_o1} and DeepSeek-R1\cite{intro_r1}. 
These models generate extensive Chains-of-Thought (CoT) \cite{intro_cot} and exhibit self-reflection, particularly in mathematical problem solving.
A pivotal training paradigm underpinning such advancements is Zero-RL \cite{intro_r1, intro_zero_rl_1, intro_zero_rl_2}, which directly applies reinforcement learning (RL) to the base model. 
This paradigm leverages on-policy rollouts and rule-based rewards to elicit and enhance innate reasoning capabilities, often outperforming supervised fine-tuning (SFT) for complex tasks. Zero-RL's success underscores its potential to unlock deeper reasoning in LLMs.

Although the Zero-RL paradigm enhances complex reasoning, we focus on two main refinements: improving learning through strategic data arrangement, and extending model capabilities beyond the confines of on-policy exploration.
\textbf{First}, although curriculum learning (CL) \cite{deng2025boostinggeneralizationreasoningvision, light_r1, related_work_cl_cl} has been widely adopted to structure learning in an easier-to-harder progression using predefined difficulty metrics, it faces a critical limitation: the model's perception of difficulty is inherently dynamic and evolves during training. 
Applying static difficulty definitions directly results in curricula that misalign with the model's real-time learning requirements, ultimately leading to suboptimal training outcomes.
\textbf{Second}, the challenge of expanding model capability boundaries presents another fundamental constraint in RL. Current approaches, including the Zero-RL paradigm, predominantly rely on on-policy methods that depend solely on self-generated rollouts. This creates an intrinsic limitation: the model's capacity for advancement becomes constrained by its pre-training knowledge base, as it lacks exposure to external reasoning patterns beyond its initial capabilities.


To address these challenges, we draw inspiration from well-established principles in cognitive science and educational psychology. First, human learning is most effective within the Zone of Proximal Development (ZPD) \cite{vygotsky1978mind}, which describes the conceptual space where tasks are challenging yet achievable with guidance. A static curriculum fails to remain within this zone as a model's competence evolves. This principle motivates our ADCL strategy, which dynamically adapts the training curriculum to consistently target the model's evolving ZPD, ensuring optimal learning conditions. Second, deep understanding of complex material is fostered not by passive absorption but through active processing, a phenomenon known as the Self-Explanation Effect \cite{chi1994eliciting}. Learners who actively explain expert solutions to themselves—rephrasing logic and connecting it to their existing knowledge—assimilate information more effectively than those who simply memorize it. This insight underpins our EGSR strategy, which guides the model to self-reformulate expert solutions, promoting a genuine integration of new reasoning patterns rather than rote imitation.

Inspired by these human learning characteristics, we propose two novel training strategies to enhance the Zero-RL training paradigm: 
\textbf{(1) Adaptive Difficulty Curriculum Learning (ADCL)} addresses the challenge of dynamic difficulty perception by employing periodic difficulty re-estimation to adjust the curriculum based on the model's difficulty perception.
\textbf{(2) Expert-Guided Self-Reformulation (EGSR)} tackles the limitations of on-policy exploration by enabling the model to learn from high-quality trajectories from expert policies. This is achieved not through mere imitation, but by guiding the model to actively reconstruct and internalize expert solutions, thereby fostering the development of new reasoning capabilities beyond its initial scope.

This paper makes the following main contributions:
\begin{itemize}
    \item {\bfseries Conceptual and Empirical Insights:} We identify the \emph{Difficulty Shift} phenomenon, which undermines static curricula, and show that Expert-Guided Self-Reformulation enables more stable, on-policy knowledge integration than direct imitation.
    \item {\bfseries Novel Strategies:} ADCL dynamically adjusts training curricula by periodically re-estimating difficulty within upcoming data batches to align with evolving model capabilities. EGSR enables knowledge assimilation by guiding models to reconstruct expert solutions within their own conceptual frameworks, rather than direct imitation.
    \item {\bfseries Comprehensive Experiments:} Our experiments confirm the efficacy of our proposed ADCL and EGSR strategies, which synergistically enhance mathematical reasoning within the Zero-RL baseline. Notably, their combined application improves performance over the standard Zero-RL baseline by 10\% on the AIME24 benchmark and 16.6\% on AIME25.
\end{itemize}

\section{Related Work}
\subsection{Curriculum Learning for LLMs}
CL is a training strategy that mimics human learning progression by systematically increasing the complexity of training data, typically following an easier-to-harder trajectory. 
This principle has demonstrated significant efficacy across various machine learning domains from computer vision (CV) \cite{related_work_cl_cv1_guo2018curriculumnet, related_work_cl_cv2_jiang2014easy} and natural language processing (NLP) \cite{related_work_cl_nlp1_platanios2019competence, related_work_cl_nlp2_tay2019simple} to reinforcement learning (RL), and notably in both the pre-training \cite{method_cl_p1_pretrain_1} and post-training \cite{method_cl_p1_posttrain_1,method_cl_p1_posttrain_2} phases of LLMs.

Specifically for reinforcement fine-tuning (RFT) of LLMs, CL strategies are increasingly being adopted to optimize the training process and improve model performance \cite{related_work_cl_llm2, related_work_cl_llm3,deng2025boostinggeneralizationreasoningvision}. Many current CL applications in this context rely on static curricula, where task difficulty is predetermined offline, and data is curated accordingly\cite{related_work_cl_llm1, related_work_cl_llm3}. Although such predefined curricula have demonstrated effectiveness, they may not dynamically adapt to the model's evolving learning state. To address this limitation of static curricula, we propose the ADCL strategy, which dynamically adjusts training curricula by periodically re-estimating difficulty within upcoming data batches to align with evolving model capabilities.

\begin{figure*}[t]
    \centering
    \includegraphics[width=1.0\linewidth]{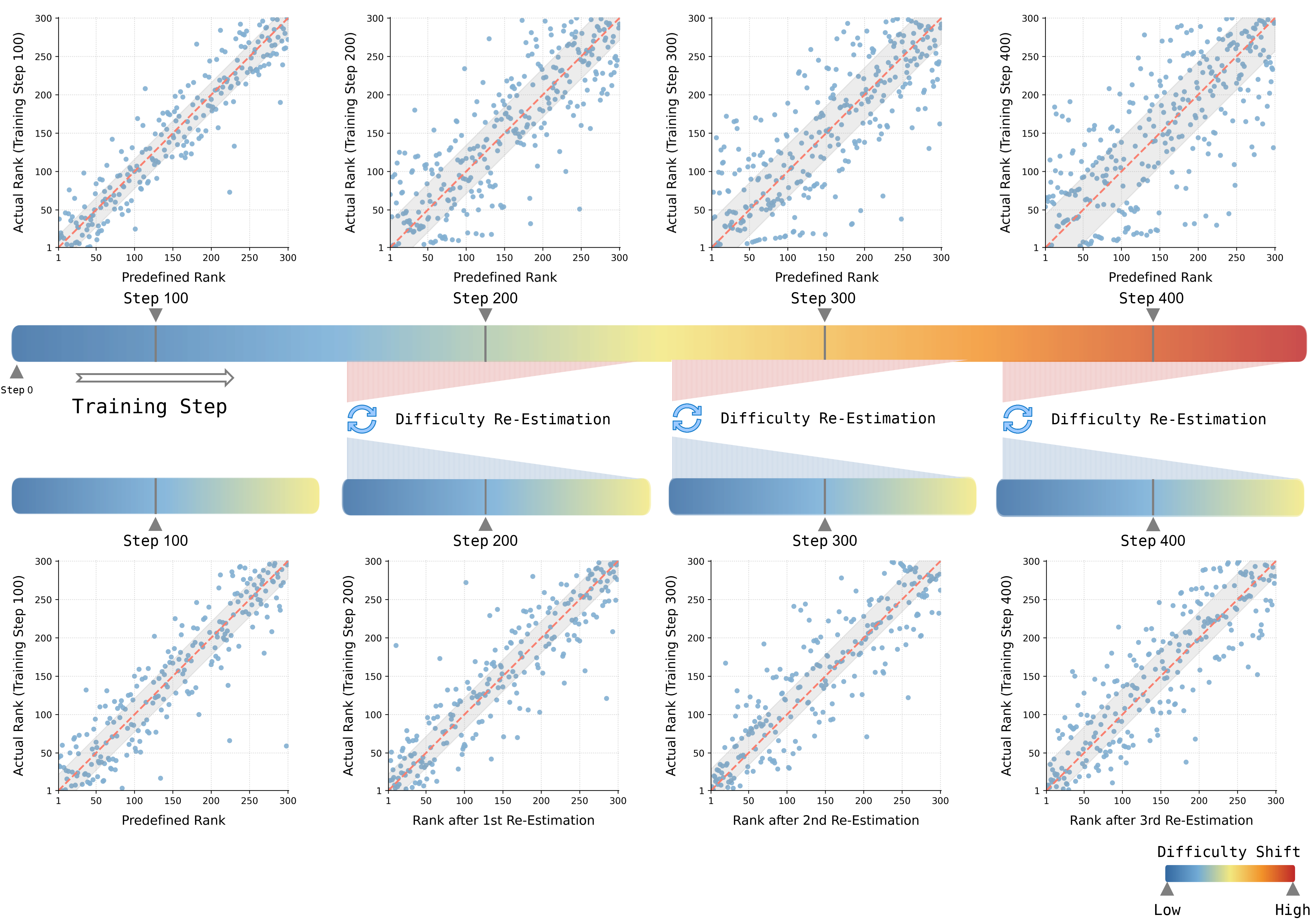}
    \caption{Illustration of the \emph{Difficulty Shift} phenomenon and the ADCL counteraction. Top Row: Scatter plots show the increasing divergence between the initial Predefined Rank (x-axis) and the model's evolving Actual Rank (y-axis) when using a predefined curriculum, demonstrating the growing \emph{Difficulty Shift} over training steps. Bottom Row: ADCL employs periodic difficulty re-estimation based on the current model state to re-sort the upcoming batch. This dynamic adjustment results in the model's Actual Ranks (y-axis) more closely aligning with the batch's re-estimated rank order (x-axis), correcting the \emph{Difficulty Shift} observed with a predefined curriculum.}
    \label{fig:difficulty_shift_exp}
\end{figure*}

\subsection{Reinforcement Learning for LLMs}
Reinforcement learning methods are broadly categorized based on data utilization into on-policy and off-policy techniques. On-policy algorithms, such as PPO \cite{related_work_rl_ppo} and TRPO \cite{related_work_trpo}, learn from data generated by the current policy, ensuring stability but often requiring extensive data. In contrast, off-policy techniques, including DQN \cite{related_work_dqn} and SAC \cite{related_work_sac}, utilize data from various policies, enhancing data efficiency but potentially introducing instability and complexity.

In the context of post-training phases, on-policy methods such as PPO and GRPO \cite{deepseek2024math} have become the de facto standard. While off-policy approaches such as DPO \cite{related_work_dpo} are also explored, a key challenge is integrating the benefits of rich off-policy data with on-policy stability. Direct integration often destabilizes training due to significant distributional differences between the learning policy and the expert policy. We propose the EGSR strategy to address this by using expert demonstrations to guide the current policy in generating improved, more on-policy trajectories.

\section{Method}
\subsection{Adaptive Difficulty Curriculum Learning}
CL is a training strategy inspired by human cognition, where models learn progressively from easier to harder examples.  
A critical challenge is the model's dynamic difficulty perception during training (a phenomenon we term \emph{Difficulty Shift}).
This dynamic shift can render the initial, fixed difficulty ranking\cite{deng2025boostinggeneralizationreasoningvision,light_r1} increasingly inaccurate, misaligning the presented curriculum with the model's real-time learning needs and potentially hindering training progression.

We provide empirical evidence of the \emph{Difficulty Shift} phenomenon in Figure \ref{fig:difficulty_shift_exp}. The color-coded progress bar (left-to-right) illustrates the standard GRPO algorithm training on a dataset. The top row illustrates training using a predefined CL approach with a predefined difficulty ordering established by the base model.
To analyze the shift, we extracted several model checkpoints during this predefined curriculum training. For each checkpoint, we assessed its current perception of difficulty by evaluating it on the immediately following 300 data points and ranking them based on the accuracy achieved by that specific checkpoint.
The resulting actual difficulty ranks (y-axis) are plotted against their original predefined ranks (x-axis, monotonically incrementing from 1 to 300) in scatter plots corresponding to each checkpoint marker, with a gray shaded region highlighting the 25th-75th percentile deviation between the actual (y) and predefined (x) ranks.
As observed in the top progression of Figure \ref{fig:difficulty_shift_exp}, training with the predefined curriculum leads to progressively increasing dispersion in the scatter plots and a widening of this quantile range, both indicating a larger \emph{Difficulty Shift} over time.
We attribute this increase in \emph{Difficulty Shift} to the increased deviation of the model from the initial model during the training process. 

To counteract this detrimental \emph{Difficulty Shift} while maintaining computational tractability, we propose Adaptive Difficulty Curriculum Learning (ADCL), a dynamic difficulty-aware training strategy. Instead of relying on a fixed, predefined order or re-evaluating the entire dataset like standard Self-Paced Learning (SPL)\cite{spl}, ADCL leverages dynamic difficulty re-estimation based on the model's current state to re-sort upcoming data batches. This mirrors how humans dynamically adjust their learning paths based on perceived task difficulty, rather than rigidly following a fixed sequence. 

The core mechanism of our ADCL algorithm is the dynamic re-estimation and re-sorting of upcoming data batches based on the model's evolving state. Specifically, the process begins by estimating difficulty scores $\delta_0(x_i)$ for all samples in dataset $\mathcal{D}$ using the initial model parameters $\theta_0$. After sorting $\mathcal{D}$ according to these scores, it is partitioned into $K$ sequential batches ${B_1, B_2, ..., B_K}$.
These batches are then processed iteratively. At iteration $k$, the model parameters $\theta_{k-1}$ are updated to $\theta_k$ by training on batch $B_k$. 
Following this, ADCL re-evaluates the difficulty scores $\delta_k(x_i)$ for elements within the subsequent batch $B_{k+1}$ using the updated model $\theta_k$. 
This batch is then internally re-sorted according to the new difficulty estimation before proceeding to the next iteration. This localized, progressive re-sorting continues until all the training is completed. The detailed pseudocode for ADCL is provided in Appendix \ref{appendix:alg_ADCL}.

As illustrated in the bottom row of Figure \ref{fig:difficulty_shift_exp}, ADCL dynamically re-estimates the difficulty of upcoming batches multiple times during training, significantly correcting the difficulty deviation compared to predefined CL. 

\begin{figure*}
    \centering
    \begin{align}
    \mathcal{J}_{\text{GRPO}}(\theta) &= \mathbb{E}{[q \sim P(Q), \{\tau_i\}_{i=1}^G \sim \pi_{\theta_{old}}(\cdot |q)]} \notag \\
    & \frac{1}{G}\sum_{i=1}^G\frac{1}{|\tau_i|} \sum_{t=1}^{|\tau_i|} \left\{ \min \left[ r_{i,t}(\theta) A_{i}, \text{clip} \left( r_{i,t}(\theta), 1 - \epsilon, 1 + \epsilon \right)  A_{i} \right] - \beta \mathbb{D}_{KL}\left[\pi_{\theta} || \pi_{ref}\right]\right\} \notag \\
    & \text{where} \quad r_{i,t}(\theta) = \frac{\pi_\theta(\tau_{i,t} | q, \tau_{i,<t})}{\pi_{\theta_{\text{old}}}(\tau_{i,t} | q, \tau_{i,<t})} \quad \text{and} \quad A_{i} = \frac{R(\tau_i)- {\rm mean}(\{R(\tau_k)\}_{k=1}^G)}{{\rm std}(\{R(\tau_k)\}_{k=1}^G) } \tag{1}\label{eq:GRPO-obj} \\[1em]
    \mathcal{J}_{\text{GRPO-EGSR}}(\theta) &= \mathbb{E}{[q,g \sim P(Q),  \{\tau_i\}_{i=1}^G \sim \pi_{\theta_{old}}(\cdot |q), \textcolor{red}{ \{\tau^{'}_i\}_{i=1}^M \sim \pi_{\theta_{old}}(\cdot |q, g) } ]} \notag \\
    & \frac{1}{G - M}\sum_{i=1}^{G- M}\frac{1}{|\tau_i|} \sum_{t=1}^{|\tau_i|} \left\{ \min \left[ r_{i,t}(\theta) A_{i}, \text{clip} \left( r_{i,t}(\theta), 1 - \epsilon, 1 + \epsilon \right)  A_{i} \right] - \beta \mathbb{D}_{KL}\left[\pi_{\theta} || \pi_{ref}\right]\right\} \notag \\
    &+ \frac{1}{M}\sum_{i=1}^{M}\frac{1}{|\tau^{'}_i|} \sum_{t=1}^{|\tau^{'}_i|} \left\{ \min  \left[ \textcolor{red}{r^{'}_{i,t}(\theta) } A_{i}, \text{clip} \left( \textcolor{red}{r^{'}_{i,t}(\theta)}, 1 - \epsilon, 1 + \epsilon \right)  A_{i} \right] - \beta \mathbb{D}_{KL}\left[\pi_{\theta} || \pi_{ref}\right]\right\} \notag \\
    & \text{where} \quad r_{i,t}(\theta) = \frac{\pi_\theta(\tau_{i,t} | q, \tau_{i,<t})}{\pi_{\theta_{\text{old}}}(\tau_{i,t} | q, \tau_{i,<t})} \quad , \quad \textcolor{red}{ r^{'}_{i,t}(\theta) = \frac{\pi_\theta(\tau_{i,t}^{'} | q, \tau_{i,<t}^{'})}{\pi_{\theta_{\text{old}}}(\tau_{i,t}^{'} | q, g, \tau_{i,<t}^{'})} } \notag \\[1em]
     & \text{and} \quad A_{i} = \frac
     {R(\tau_i)- {\rm mean}( \textcolor{red}{ \{R(\tau_k)\}_{k=1}^{G-M} + \{R(\tau^{'}_k)\}_{k=1}^{M})}) }
     {{\rm std}(\textcolor{red}{ \{R(\tau_k)\}_{k=1}^{G-M} + \{R(\tau^{'}_k)\}_{k=1}^{M} }) }  
     \tag{2} \label{eq:GRPO-EGSR}
    \end{align}
\end{figure*}

\subsection{Expert-Guided Self-Reformulation}
RL for LLMs, particularly in reasoning tasks, often functions as a process to enhance sample efficiency. Although RL can improve accuracy on problems the model can occasionally solve, it fundamentally struggles to elicit solutions for problems entirely outside the model's current capabilities. This limitation implies that the base model's inherent knowledge restricts the upper bound of performance achievable through standard on-policy RL techniques\citep{yue2025doesreinforcementlearningreally}.

The standard objective for certain on-policy RL algorithms, such as GRPO detailed in Eq.\ref{eq:GRPO-obj}, aims to optimize the policy $\pi_\theta$ using $G$ trajectories $\{\tau_i\}_{i=1}^G$ sampled from a previous policy $\pi_{\theta_{\text{old}}}$ for a given question $q$. 
However, a frequent challenge in practical training scenarios with such algorithms is the "zero-reward" situation. This occurs when all $G$ rollouts from $\pi_{\theta_{\text{old}}}$ yield a reward of zero, i.e., $R(\tau_i) = 0$ for all $i$. 
This outcome primarily stems from problem complexity exceeding the model's current capabilities.
In such zero-reward scenarios, advantage estimates $\hat{A}_{i,t}$ across all actions effectively vanish, resulting in null gradient updates that fail to contribute to policy improvement. 
This learning impasse highlights the need to incorporate external guidance from an off-policy expert policy $\pi_\phi$. Such guidance, often provided through demonstrations generated by this policy, can introduce meaningful learning signals and potentially infuse the model with knowledge beyond its current capabilities.

One straightforward way to leverage guidance from this expert policy is to replace a portion of on-policy trajectories with $M$ expert demonstrations\cite{hester2018deep, liu2022improved}, creating a hybrid trajectory set $\mathcal{T}_{\text{mixed}} = \mathcal{T}_{\text{on}} \cup \mathcal{T}_{\text{off}}$, where $\mathcal{T}_{\text{on}} = \{\tau_i \mid \tau_i \sim \pi_{\theta_\text{old}}(\cdot|q), i=1,...,G-M\}$ and $ \mathcal{T}_{\text{off}} = \{\tau_j \mid \tau_j \sim \pi_{\phi}(\cdot|q), j=1,...M \}$.
When incorporating off-policy demonstrations, importance sampling must be applied to correct the distribution shift, modifying the probability ratio in the GRPO objective as:
\begin{equation}
    r_{i,t}(\theta) = 
    \begin{cases}
        \frac{\pi_\theta(\tau_{i,t} | q, \tau_{i,<t})}{\pi_{\theta_{\text{old}}}(\tau_{i,t} | q, \tau_{i,<t})}, & \text{if } \tau_i \in \mathcal{T}_{\text{on}} \\[8pt]
        \frac{\pi_\theta(\tau_{i,t} | q, \tau_{i,<t})}{\pi_{\phi}(\tau_{i,t} | q, \tau_{i,<t})}, & \text{if } \tau_i \in \mathcal{T}_{\text{off}}
    \end{cases}
    \tag{3}
\end{equation}
And the advantage estimates are computed as:
\begin{equation}
    A_{i} = \frac
    {R(\tau_i)- {\rm mean}(R(\tau_{\text{mixed}}))}
    {{\rm std}(R(\tau_{\text{mixed}}))}
    \tag{4}
\end{equation}

While this approach is theoretically sound, in practice the expert policy $\pi_\phi$ is typically inaccessible or uses incompatible tokenization schemes, making the probability ratio calculation infeasible. Even when $\pi_\phi$ is available, the substantial distributional differences between policies often produce extreme importance weights with destabilizing variance. Our experiments in Section \ref{sec:exp:main_results} demonstrate that incorrectly treating expert demonstrations as on-policy samples may degrade performance on complex reasoning tasks.

We argue that the core issue lies in the distributional mismatch between the reference solution trajectories generated by an external expert policy $\pi_{\phi}$ and the trajectories naturally produced by the current policy $\pi_{\theta_\text{old}}$. 
Inspired by the human learning analogy of restating solutions, we propose a method that circumvents the need to estimate $\pi_\phi$ or handle large importance weights. Instead of directly incorporating the off-policy trajectory $\tau_j \sim \pi_\phi$, we use expert demonstrations to guide $\pi_{\theta_{\text{old}}}$ in generating more effective trajectories $\tau^{'}_{i} \sim \pi_{\theta_{\text{old}}}(\cdot | q, g)$, where $g$ represents the guidance information. Similar to how students learn by rephrasing solutions in their own words, this approach produces trajectories that are naturally aligned with the model's current capabilities while incorporating expert knowledge. This guided generation creates samples that remain fundamentally "on-policy" while bridging the gap toward expert performance.

\begin{figure}[t]
    \centering
    \includegraphics[width=1.0\linewidth]{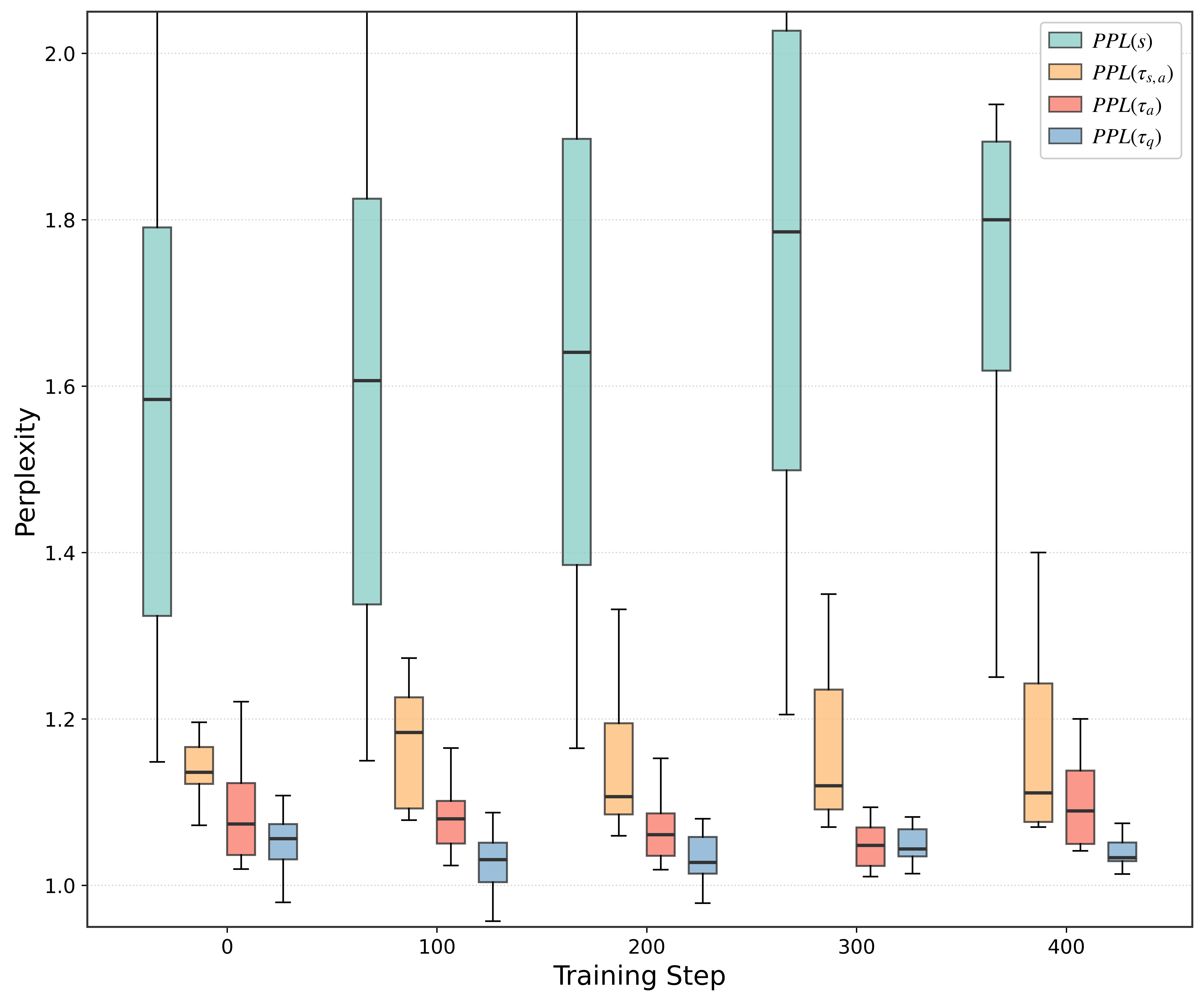}
    \caption{Perplexity (PPL) of four trajectory types under the current model policy $\pi_\theta$ across training checkpoints: unguided model generations $\boldsymbol{\tau_q}$ (blue), reference expert solutions $\boldsymbol{s}$ (green), model-generated trajectories guided by $(s,a)$ $\boldsymbol{\tau_{s,a}}$ (orange), and those guided by $a$ alone $\boldsymbol{\tau_a}$ (red). Lower PPL indicates better alignment with $\pi_\theta$.}
    \label{fig:ppl}
\end{figure}

To verify our hypothesis, we experimentally measured the model's perplexity (PPL) on four distinct trajectory types across multiple training checkpoints, using 300 problems defined by $(q,s,a)$ triples (question, expert solution, answer); these results are shown in Figure \ref{fig:ppl}. 
We evaluated PPL for: (1) unguided model generations $\boldsymbol{\tau_q}$, (2) reference expert solutions $\boldsymbol{s}$, (3) model-generated trajectories $\boldsymbol{\tau_{s,a}}$ guided by both $s$ and $a$, and (4) trajectories $\boldsymbol{\tau_a}$ guided by $a$ alone. 
Throughout training, the PPL of reference expert solutions $s$ consistently remained the highest, while the PPL of the model's unguided outputs $\tau_q$  was the lowest, serving as a baseline for its natural generative distribution. Crucially, the PPL of $\tau_{s,a}$ and $\tau_{a}$ was found to be substantially lower than $PPL(s)$ and markedly closer to $PPL(\tau_q)$. These PPL results indicate that guiding the model to generate solutions yields trajectories significantly more aligned with its current policy (i.e., more 'on-policy') than directly using off-policy expert solutions.

Building on our PPL analysis (Figure \ref{fig:ppl}), we propose Expert-Guided Self-Reformulation (EGSR), a novel reinforcement learning training strategy designed to enhance on-policy reinforcement learning algorithms, like GRPO. The EGSR training objective, presented in Equation.\ref{eq:GRPO-EGSR}, integrates $M$ expert-guided trajectories $\tau'_i \sim \pi_{\theta_{old}}(\cdot|q,g)$ (where $g$ is guidance derived from expert policy $\pi_\phi$) with $G-M$ standard rollouts, particularly when initial exploration yields zero rewards. For these guided trajectories, the ratio $r'_{i,t}(\theta)$ is calculated with respect to their specific generation policy $\pi_{\theta_{old}}(\cdot|q,g)$. This is justified as our PPL findings indicate that trajectories from $\pi_{\theta_{old}}(\cdot|q,g)$ are distributionally closer to the model's unguided outputs than trajectories from the original expert policy $\pi_\phi$, thereby promoting more stable, near-on-policy learning while still leveraging expert knowledge. The detailed pseudocode for EGSR is provided in Appendix \ref{appendix:alg_EGSR}.

\begin{table*}[h]
\centering
\begin{tabular}{lcccccc}
\toprule
\textbf{Model} & \textbf{MATH} & \textbf{AIME24} & \textbf{AIME25} & \textbf{AMC23} & \textbf{Minervamath}  \\\midrule
Qwen2.5-7B                  & 69.40 & 16.67 & 16.67 & 32.50 & 15.44             \\
 + SFT                      & 70.80 & 13.33 & 13.33 & 35.00 & 16.91              \\
 + RL                       & 72.40 & 26.67 & 16.67 & 45.00 & 18.01             \\\midrule
 + PCL                      & 75.40 & 30.00 & 23.33 & 50.00 & 22.42              \\
 + ADCL                     & 76.20 & \underline{33.33} & \underline{30.00} &  \underline{55.00} & 22.78 &              \\\midrule
 + off-policy              & 66.00 & 23.33 & 16.67 & 30.00 & 18.65              \\
 + EGSR($a$)                & 72.60 & 30.00 & 20.00 & 50.00 & 20.96             \\
 + EGSR($s,a$)              & \underline{79.40} & \underline{33.33} & \underline{30.00} & \textbf{57.50} & \underline{24.63}           \\\midrule
 + ADCL \& EGSR             & \textbf{81.80} & \textbf{36.67} & \textbf{33.33} & \underline{55.00} & \textbf{25.74} \\
\bottomrule
\end{tabular}
\caption{Performance of Qwen2.5-7B with various training strategies on mathematical reasoning benchmarks, \texttt{+SFT}: Supervised Fine-Tuning, \texttt{+RL}: Standard GRPO algorithm, \texttt{+PCL}: GRPO with Predefined Curriculum Learning, \texttt{+ADCL}: GRPO with our Adaptive Difficulty Curriculum Learning, \texttt{+off-policy}: GRPO with direct replacement of rollouts by expert solutions, \texttt{+EGSR($a$)}: Our ESGR method using only expert answer ($a$) for guidance, \texttt{+EGSR($s,a$)}: Our ESGR method using expert solution ($s$) and answer ($a$) for guidance, \texttt{+ADCL\&EGSR}:  Combination of our ADCL and EGSR($s,a$) methods. Bold and underline represent the 1st and 2nd in performance.}
\label{tab:main_results}
\end{table*}

\section{Experiments}
\label{sec:exp}
\subsection{Setup}
\textbf{(1) Datasets:} Our datasets are curated from high-quality reasoning corpora including S1 \cite{exp_train_dataset_s1} and DeepScaleR \cite{exp_train_dataset_deepscalar}. We applied filtering criteria to ensure solutions are verifiable by Math-Verify\footnote{https://github.com/huggingface/Math-Verify} and problems require computational rather than proof-based solutions.
We estimated problem difficulty using Qwen2.5-7B\cite{qwen2} as our base model, performing 32 rollouts per problem to calculate accuracy rates. From this evaluation, we created a collection of 6,894 problems with accuracy rates between 10\%-90\%, termed \textbf{BaseSet-7K}. To provide a more extensive set of instances for our EGSR method to leverage for guidance, we supplemented BaseSet-7K by incorporating an additional 3,000 problems where the base model consistently achieved 0\% accuracy. This formed the augmented collection termed \textbf{AugSet-10K}.

\noindent \textbf{(2) Benchmarks:} To evaluate our models' mathematical reasoning capabilities, we use five established benchmarks: MATH500\cite{benchmark_math_500}, AIME24\footnote{https://huggingface.co/datasets/hendrydong/aime24}, AIME25\footnote{https://huggingface.co/datasets/TIGER-Lab/AIME25}, AMC23\footnote{https://huggingface.co/datasets/zwhe99/amc23}, and Minervamath\cite{benchmark_minevarmath}. We report pass@8 for AIME24 and AIME25 to mitigate the high variance that would result from pass@1 on these smaller benchmark sets, and use standard pass@1 for all other benchmarks.

\noindent \textbf{(3) Settings:} We implemented our training framework using TRL\cite{vonwerra2022trl} with Qwen2.5-7B as the base model. For GRPO, we used 8 rollouts per problem, a global batch size of 1024, and a fixed learning rate of $1 \times 10^{-6}$. Our reward function, drawing inspiration from \cite{deepseek2024math}, is a composite reward: $R(\tau) = \lambda_1 \cdot R_{\text{format}}(\tau) + \lambda_2 \cdot R_{\text{accuracy}}(\tau)$,
where
$$R_{\text{format}}(\tau) = \begin{cases} 1 & \text{if } \tau \text{ follows the output format} \\ 0 & \text{otherwise} \end{cases}$$
and
$$R_{\text{accuracy}}(\tau) = \begin{cases} 1 & \text{if } \tau \text{ contains the correct answer} \\ 0 & \text{otherwise} \end{cases}$$, we set $\lambda_1 = 1.0$ and $\lambda_2 = 2.0$.

For the ADCL strategy, we utilized the BaseSet-7K dataset, and our implementation divided the curriculum into four difficulty-based batches and performed three difficulty re-estimations.

For the EGSR strategy, we utilized the AugSet-10K dataset, and explored two guidance strategies: using only the expert's final answer $a$ (EGSR($a$)) or using both the answer $a$ and the solution process $s$ (EGSR($s,a$)). Considering that GRPO's learning effectiveness is maximized when the success rate is around 50\% \cite{success_50_rate}, we thus generated 4 guided trajectories in such instances. Complete hyperparameters and training results are provided in Appendix \ref{appendix:hyperparams}.

\subsection{Main Results}
\label{sec:exp:main_results}
Our main results are presented in Table \ref{tab:main_results}. 
Initially, we observed that while SFT (+SFT) showed slight improvements on some benchmarks, its performance on more complex benchmarks such as AIME24 and AIME25 actually decreased compared to the base model. In contrast, the standard RL (+RL) method brought more significant and stable performance improvements on most benchmarks. These findings indicate that RL adapts to complex tasks better than SFT.

Comparing CL strategies, our proposed Adaptive Difficulty Curriculum Learning (+ADCL) consistently outperforms Predefined Curriculum Learning (+PCL).
This advantage stems from ADCL's curriculum design, which ensures more effective alignment with the model's evolving state.

Turning to expert guidance, the naive incorporation of expert solutions via direct off-policy guidance (+off-policy) led to a degradation in performance compared to the +RL baseline.
In contrast, our EGSR strategies, both EGSR(a) and EGSR(s,a), surpassed this naive approach, with EGSR(s,a) achieving the most favorable results among the guidance methods.
While EGSR(a), which uses only the final expert answer for guidance, we observed instances where it could lead the model to generate trajectories that reach the correct answer via a flawed or incomplete reasoning process, even though such trajectories might exhibit lower perplexity. 
An example illustrating such a scenario, where the model correctly predicts the answer but with an erroneous solution when guided by the answer alone, is provided in Appendix \ref{appendix:case_study}.
EGSR(s,a), by incorporating guidance from both the expert's solution process and the final answer, encourages the model to reformulate a more complete and coherent reasoning path within its own conceptual framework, leading to more effective improvements in reasoning capabilities.

Furthermore, our two proposed training strategies, ADCL and EGSR, can be synergistically combined for even better performance. When used together (+ADCL \& EGSR), these strategies show substantial gains over the standard RL approach. Specifically, the combined approach achieved a 10\% improvement on AIME24, a 16.66\% improvement on AIME25, and a 7.73\% improvement on Minervamath compared to the +RL baseline.

\section{Analysis}
\subsection{Generalization to Other Architectures}

To assess the generalizability of our proposed strategies, we conducted additional experiments on models with different architectures and scales: Llama3.1-8B-Instruct\cite{grattafiori2024llama3herdmodels} and Qwen2.5-1.5B\cite{qwen2}. As shown in Table \ref{tab:generalization_results}, the combination of ADCL and EGSR consistently and significantly outperforms the RL baseline across both models. These results confirm that the benefits of our methods are not specific to a single model family but are broadly applicable, highlighting their robustness.


\begin{table}[!h]
\small
\centering
\begin{tabularx}{\columnwidth}{l *{4}{>{\centering\arraybackslash}X}}
\toprule
\textbf{Model} & \textbf{MATH} & \textbf{AIME24} & \textbf{AIME25} & \textbf{AMC23} \\
\midrule
Qwen2.5-1.5B   & 32.60 & 10.00 & 3.33 & 22.50 \\
+SFT           & 35.80 & 10.00 & 0.00 & 27.50 \\
+RL            & 59.00 & 16.67 & 6.67 & 35.00 \\
+PCL           & 61.20 & 16.67 & 10.00 & 35.00 \\
+ADCL          & \underline{67.00} & 16.67 & 10.00 & 40.00 \\
+off-policy    & 30.80 & 6.67 & 0.00 & 7.50 \\
+EGSR          & 64.60 & \textbf{26.67} & \underline{16.67} & \underline{45.00} \\
+ADCL \& EGSR     & \textbf{69.80} & \underline{23.33} & \textbf{20.00} & \textbf{47.50} \\
\bottomrule
\end{tabularx}

\vspace{1em} 

\begin{tabularx}{\columnwidth}{l *{4}{>{\centering\arraybackslash}X}}
\toprule
\textbf{Model} & \textbf{MATH} & \textbf{AIME24} & \textbf{AIME25} & \textbf{AMC23} \\
\midrule
Llama3.1-8B    & 46.20 & 13.33 & 3.33 & 27.50 \\
+SFT           & 44.00 & 6.67 & 6.67 & 22.50 \\
+RL            & 53.20 & 13.33 & 6.67 & 32.50 \\
+PCL           & 54.60 & \underline{20.00} & \underline{10.00} & 35.00 \\
+ADCL          & 56.40 & \textbf{23.33} & \underline{10.00} & \underline{40.00} \\
+off-policy    & 32.00 & 13.33 & 6.67 & 15.00 \\
+EGSR          & \underline{60.40} & \underline{20.00} & \textbf{13.33} & \underline{40.00} \\
+ADCL \& EGSR     & \textbf{60.60} & \textbf{23.33} & \textbf{13.33} & \textbf{42.50} \\
\bottomrule
\end{tabularx}
\caption{Performance on Qwen2.5-1.5B and Llama3.1-8B-Instruct, using the same experimental setup as in Table \ref{tab:main_results}. Bold and underline represent the 1st and 2nd in performance.}
\label{tab:generalization_results}
\end{table}

\subsection{ADCL Training Dynamic}
\begin{figure}[t]
    \centering
    \includegraphics[width=1.0\linewidth]{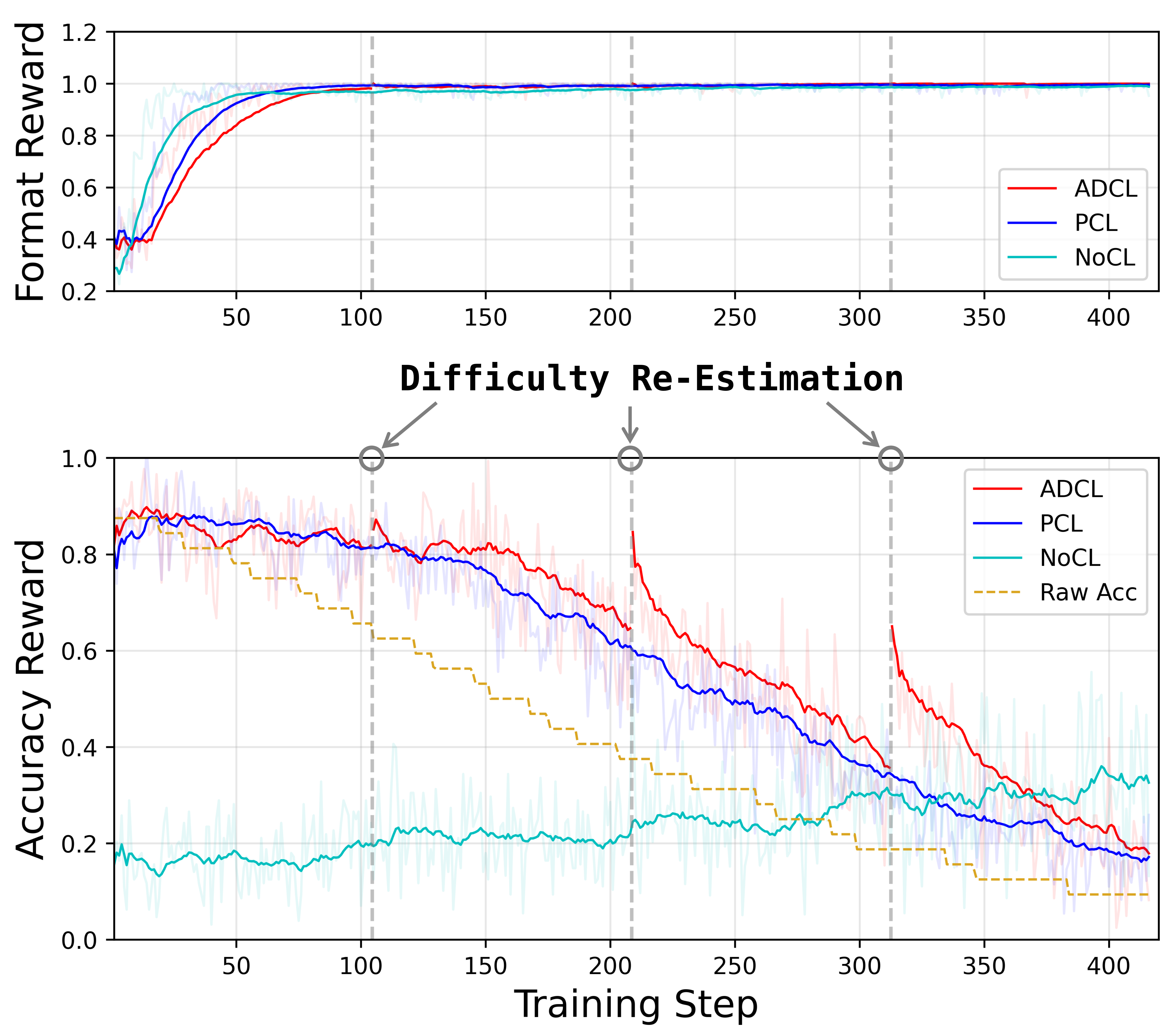}
    \caption{Comparison of reward dynamics under different training strategies. The curves represent Adaptive Difficulty Curriculum Learning (ADCL, red), Predefined Curriculum Learning (PCL, blue), and No Curriculum learning (NoCL, cyan). The dashed yellow line (Raw Acc) indicates the basemodel's performance on curriculum-ordered data. }
    \label{fig:exp_1}
\end{figure}
Figure \ref{fig:exp_1} illustrates the ADCL training dynamics under different strategies. Standard RL without curriculum (NoCL) shows a steady increase in accuracy reward. In contrast, curriculum-based strategies (ADCL and PCL), despite showing a decrease in accuracy reward later in training due to increased problem difficulty, consistently maintain higher rewards compared to the baseline raw accuracy at corresponding difficulty levels. This confirms that learning is occurring and the reward decrease is primarily driven by the curriculum's progression to harder tasks. The key advantage of our proposed ADCL over PCL lies in its adaptive nature. PCL's reward curve declines relatively smoothly because its static difficulty ranking suffers from \emph{Difficulty Shift}; training batches inevitably mix samples of varying actual difficulty relative to the model's current state, averaging out the perceived challenge. ADCL, however, employs periodic difficulty re-estimation and batch re-sorting based on the current model's perception. Appendix \ref{appendix:qadcl} quantifies this re-sorting against the \emph{Difficulty Shift}. This leads to a steeper reward curve, particularly after re-estimation points, because ADCL immediately presents the model with a batch ordered according to its updated sense of "easy-to-hard". 

\subsection{Impact of Expert Guidance on Capability Boundaries}

\begin{table}[!h]
\small
\centering
\setlength{\tabcolsep}{1pt}
\begin{tabular}{lcccccc}
\toprule
\textbf{Model} & \textbf{MATH} & \textbf{AIME25} & \textbf{AMC23} & \textbf{Minervamath}  \\\midrule
Qwen2.5-7B                      & 89.40 & 30.00 & 87.50 & 49.63           \\
 + SFT                          & 90.20 & 23.33 & \underline{90.00} & 50.36           \\
 + RL                           & 88.60 & 26.67 & 85.00 & 49.26            \\
 + off-policy          & \underline{93.20} & \underline{40.00} & \textbf{92.50} & \underline{51.47}           \\
 + EGSR($s,a$)                  & \textbf{93.40} & \textbf{46.67} & \textbf{92.50} & \textbf{51.84}            \\
\bottomrule
\end{tabular}
\caption{Model performance comparison using pass@32 metric, which better reflects capability boundaries.}
\label{tab:pass@32}
\end{table}

To further investigate whether expert guidance genuinely expands a model's intrinsic problem-solving capabilities, we evaluated various methods using a pass@32 metric. Compared to pass@1 or pass@8, pass@32 allows the model substantially more opportunities to explore diverse solution paths, offering a clearer view of its underlying capability boundaries. The results presented in Table \ref{tab:pass@32} reveal that standard RL, while effective at optimizing for common success, did not expand and even slightly contracted the model's capability boundary compared to the base model. SFT provided a marginal improvement. In contrast, methods incorporating expert guidance demonstrated more significant expansions of these boundaries. Notably, our EGSR(s,a) strategy achieved the largest improvement, increasing the pass@32 accuracy on AIME25 by a substantial 16.67\% over the base model, suggesting that it fosters deeper knowledge assimilation rather than merely refining existing skills. 

\section{Conclusion}
Drawing inspiration from human learning strategies, this work addresses key challenges in enhancing the capabilities of large language models to solve complex tasks. We first observed the \emph{Difficulty Shift} phenomenon, where a model's perception of problem difficulty changes dynamically during training, hindering static curriculum learning. To counteract this, we propose Adaptive Difficulty Curriculum Learning (ADCL), which periodically re-estimates difficulty to maintain an aligned learning path.
Secondly, recognizing that existing methods often fail to help models assimilate knowledge beyond their initial capabilities, we introduce Expert-Guided Self-Reformulation (EGSR). EGSR guides models to actively reformulate expert solutions within their own conceptual framework, rather than relying on direct imitation, fostering deeper knowledge assimilation.
Experiments validate our proposed ADCL and EGSR, showing they significantly outperform baselines, especially in combination. These findings highlight the value of incorporating adaptive human-like learning mechanisms into LLM training.

\section*{Limitations}
While we identify and address the \emph{Difficulty Shift} phenomenon within our RL framework, a broader investigation into its prevalence and characteristics across different training paradigms, such as Supervised Fine-Tuning (SFT), and various domains would offer a more comprehensive understanding. The current implementation of ADCL employs a fixed number of difficulty re-estimations and re-sortings; an exploration of varying these frequencies could further illuminate its impact on correcting the \emph{Difficulty Shift} and optimizing performance. The efficacy of EGSR is closely tied to the quality of expert demonstrations, and while our study assumes access to high-quality demonstrations, the precise impact of imperfections, errors, or biases within these demonstrations on the model's learning process warrants further investigation. Our findings are primarily demonstrated using a specific model architecture, size, and datasets focused on mathematical reasoning; further studies would be beneficial to ascertain the generalizability of our proposed ADCL and EGSR strategies across a wider range of model types, scales, and diverse application domains.

\bibliography{custom}

\iftrue
\clearpage
\appendix
\onecolumn
\section{Pseudocode for ADCL}
\label{appendix:alg_ADCL}
\begin{algorithm}
\caption{Adaptive Difficulty Curriculum Learning}
\label{alg:ADCL}
\textbf{Input:} Dataset $\mathcal{D} = \{x_i\}_{i=1}^{M}$; initial policy model $\pi_{\theta_0}$; number of batches $K$;. \\
\begin{algorithmic}[1]
    \State $\forall x_i \in \mathcal{D}$, compute difficulty score $\delta_0(x_i) = f(\pi_{\theta_0}, x_i)$ \Comment{Initial difficulty estimation}
    \State $\mathcal{D}_{sorted} \leftarrow \text{Sort}(\mathcal{D}, \delta_0)$ \Comment{Sort based on increasing difficulty}
    \State Partition $\mathcal{D}_{sorted}$ into $\{B_1, B_2, \ldots, B_K\}$ where $|B_k| = \lfloor M/K \rfloor$ or $\lceil M/K \rceil$ \label{line:partition}
    \For{$k = 1$ to $K$} \label{line:main_loop}
        \State $\theta_k \leftarrow \text{TrainModel}(\theta_{k-1}, B_k)$  \Comment{Update $\pi_{\theta}$ using RL algorithm (e.g., GRPO)}
        \If{$k < K$}
            \State $\forall x_i \in B_{k+1}$, compute $\delta_k(x_i) = f(\pi_{\theta_k}, x_i)$ \Comment{Re-estimate difficulty}
            \State $B_{k+1} \leftarrow \text{Sort}(B_{k+1}, \delta_k)$  \Comment{Re-sort next batch}
        \EndIf
    \EndFor
\end{algorithmic}
\textbf{Output:} $\pi_{\theta_K}$.
\end{algorithm}

\section{Pseudocode for EGSR}
\label{appendix:alg_EGSR}
\begin{algorithm}
\caption{Expert-Guided Self-Reformulation}
\label{alg:EGSR}
  \textbf{Input} initial policy model $\pi_{\theta_{\text{init}}}$; reward models $r_\phi$; Dataset $\mathcal{D} = \{(q_i, s_i, a_i)\}_{i=1}^{N}$, where $q_i$ is the question, $s_i$ is the solution, and $a_i$ is the answer;
  \begin{algorithmic}[1]
    \State policy model $\pi_\theta \leftarrow \pi_{\theta_{\text{init}}}$
    \For{iteration = 1, \dots, I}
      \For{step = 1, \dots, M}
        \State Sample a batch $\mathcal{D}_{b}$ from $\mathcal{D}$
        \State Update the old policy model $\pi_{\theta_{old}} \leftarrow \pi_{\theta}$ 
        \State Sample $G$ trajectories $\{\tau_i\}_{i=1}^G \sim \pi_{\theta_{old}} (\cdot \mid q) $  for each question $q \in \mathcal{D}_b$
        \State Compute rewards $\{R(\tau_i)\}_{i=1}^{G}$ for each sampled trajectory $\tau_i$ using $r_{\phi}$ 
        \If{$\sum_{i=1}^{G} R(\tau_i) = 0$}  \Comment{All rewards are zero}
        \State Generate guided trajectories $\{\tau_{i}^{'}\}_{i=1}^{M} \sim \pi_{\theta_{old}} (\cdot \mid f(q, s, a))$ 
        \State Create mixed trajectory $\tau_{\text{mixed}} = \{ \tau_k \}^{G-M}_{k = 1}  \cup \{ \tau_k^{'} \}^{M}_{k = 1}  $
        \State Recompute rewards for $\tau_{\text{mixed}}$
      \EndIf
      \State Compute advantages $A_{i,t}$ for each token $t$ in each trajectory $\tau_i$ from $\mathcal{T}_{\text{mixed}}$
      \For{GRPO iteration = 1, \dots, $\mu$}
        \State Update $\pi_{\theta}$ by maximizing the GRPO-EGSR objective in Eq. \ref{eq:GRPO-EGSR}
        \EndFor
      \EndFor
    \EndFor 
  \end{algorithmic}
  \textbf{Output} $\pi_\theta$
\end{algorithm}
\clearpage
\twocolumn
\section{Training Hyperparameters}
\label{appendix:hyperparams}
The complete hyperparameters used during training are shown in Table \ref{tab:hyperparams}.
\begin{table}[H]
\setlength{\tabcolsep}{1.5pt}
\centering
\begin{tabular}{ll}
\toprule
\multicolumn{2}{c}{\textbf{Training Setup}} \\ \midrule
global\_batch\_size & 1024 \\
per\_device\_batch\_size & 8 \\
num\_machines & 4 \\
num\_devices & 32 (4 × 8 H20s) \\
\midrule
\multicolumn{2}{c}{\textbf{Rollout Configuration}} \\ \midrule
rollout\_backend & vllm \\
tensor\_parallel\_size & 1 \\
data\_parallel\_size & 8 \\
num\_generation & 8 \\
max\_completion\_length & 4096 \\
temperature & 0.7 \\
\midrule
\multicolumn{2}{c}{\textbf{GRPO Training Configuration}} \\ \midrule
learning rate & 1e-6 (constant) \\
beta & 0 \\
epsilon & 0.2 \\
reward functions & format, accuracy  \\
reward weights & 1.0, 2.0 \\
num\_iteration & 4 \\
gradient\_accumulation\_steps & 4 \\
\midrule
\multicolumn{2}{c}{\textbf{SFT Training Configuration}} \\ \midrule
learning rate & 2e-5 (decay) \\
max\_length & 8192 \\
gradient\_accumulation\_steps & 1 \\
\bottomrule
\end{tabular}
\caption{Training Hyperparameters}
\label{tab:hyperparams}
\end{table}

Detailed training curves can be found in Figure \ref{fig:sub_adcl}, Figure \ref{fig:sub_egsr}, and Figure \ref{fig:sub_sft}.

\begin{figure}[h]
    \centering
    \includegraphics[width=1.0\linewidth]{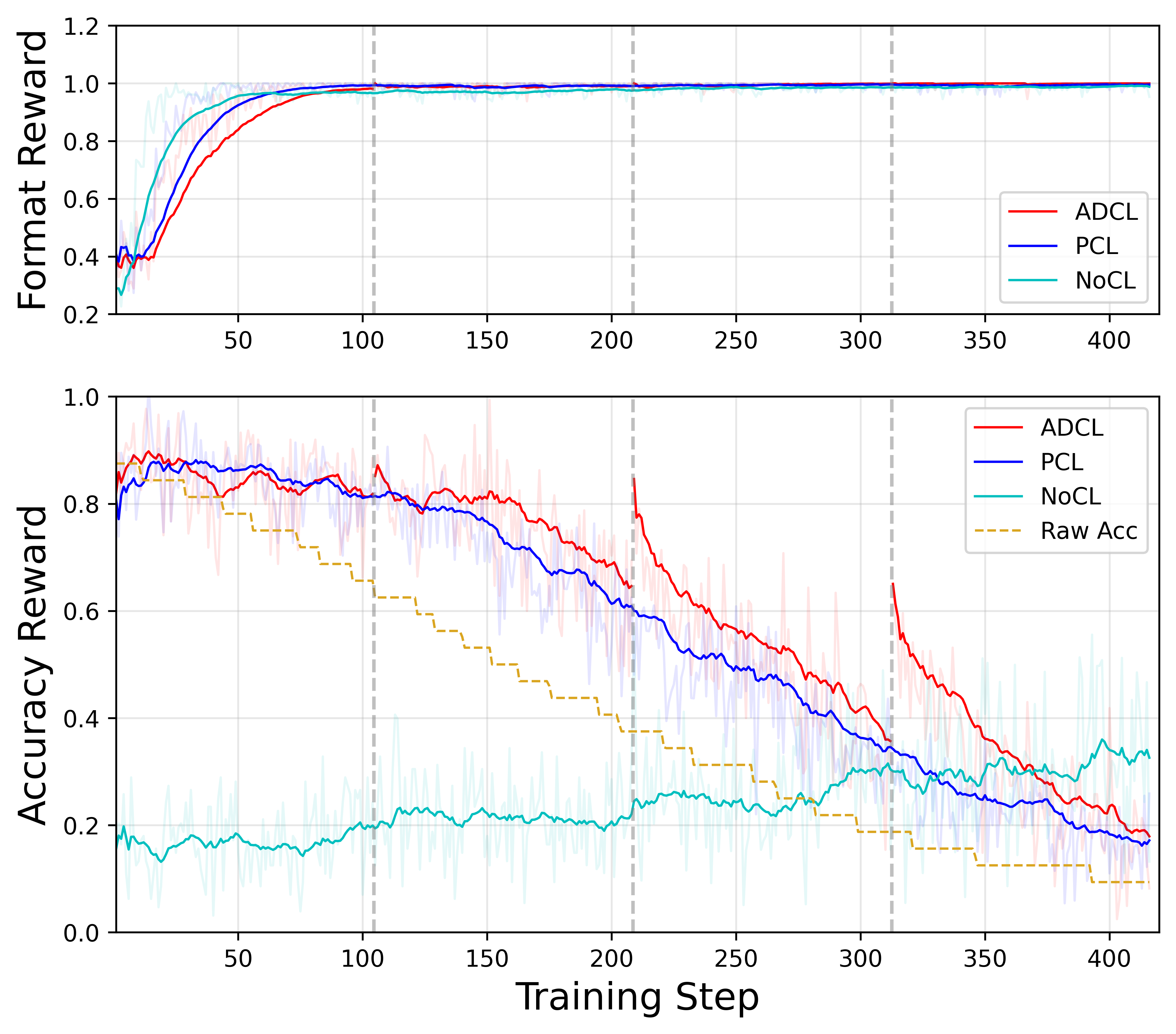}
    \caption{Reward curves from training the ADCL strategy and relevant baseline methods}
    \label{fig:sub_adcl}
\end{figure}

\begin{figure}[h]
    \centering
    \includegraphics[width=1.0\linewidth]{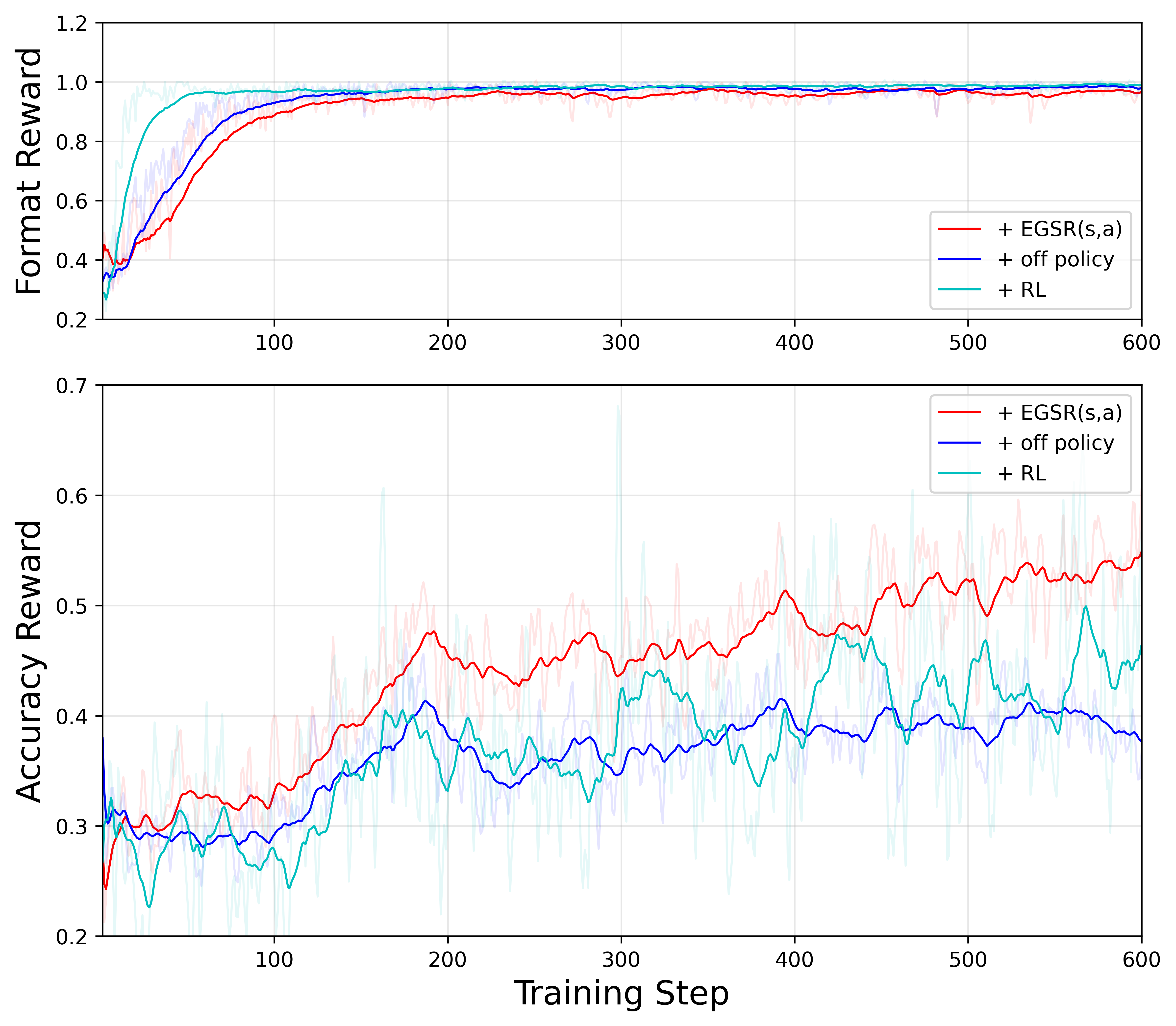}
    \caption{Reward curves from training the EGSR strategy and relevant baseline methods}
    \label{fig:sub_egsr}
\end{figure}

\begin{figure}[h]
    \centering
    \includegraphics[width=1.0\linewidth]{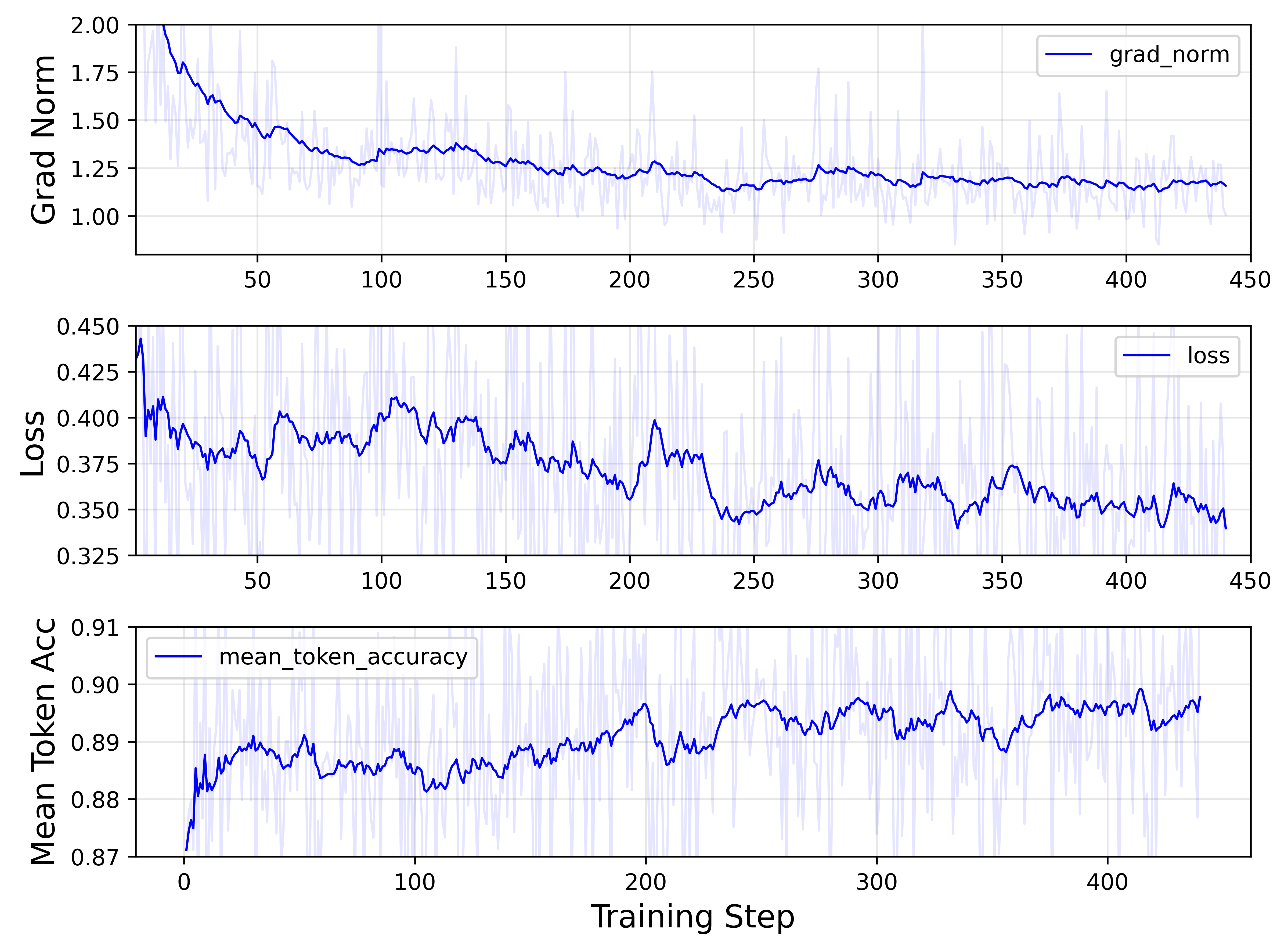}
    \caption{Supervised Fine-Tuning (SFT) training curve}
    \label{fig:sub_sft}
\end{figure}

\clearpage
\section{Quantifying ADCL Re-sorting Dynamics}
\label{appendix:qadcl}
\begin{table}[!h]
\small
\centering
\begin{tabular}{@{}lc@{}}
\toprule
\textbf{Curriculum Batch} & \textbf{NIR} \\
\midrule
Batch 1 (No Re-estimation) & 0.000 \\
Batch 2 (After 1st Re-estimation)  & 0.331 \\
Batch 3 (After 2nd Re-estimation)  & 0.363 \\
Batch 4 (After 3rd Re-estimation)  & 0.369 \\
\bottomrule
\end{tabular}
\caption{Normalized inversion rates for the subsequent batch, quantifying the re-sorting difference after each ADCL difficulty re-estimation event.}
\label{tab:adcl_inversions}
\end{table}
To verify ADCL's re-sorting mechanism, we measured the disagreement between the predefined and the dynamically re-estimated sample orders within batches using the Normalized Inversion Rate (NIR). This metric calculates the fraction of sample pairs whose relative order is flipped (0=identical, higher=more different). Table~\ref{tab:adcl_inversions} shows the NIR measured for each curriculum batch after the preceding difficulty re-estimation event. The rate increased from 0 for the initial batch (Batch 1, where no re-estimation was applied) to 0.331, 0.363, and 0.369 for Batches 2, 3, and 4 (processed after the 1st, 2nd, and 3rd re-estimations, respectively). These substantial, non-zero rates confirm re-sorting occurs (affecting 33-37\% of pairs). This increasing NIR trend indicates that the underlying \emph{Difficulty Shift} becomes more pronounced as training progresses, necessitating greater adaptation by ADCL.

\section{Prompts for EGSR}
\begin{tcolorbox}[
    breakable, 
    colback=white!95!gray,
    colframe=gray!50!black,
    rounded corners,
    label={prompt-dot}, 
    title={Prompt for EGSR($a$): Generating Trajectory $\tau(a)$},
    before upper={\small}
]
\begin{lstlisting}[breaklines=true, xleftmargin=0pt, breakindent=0pt, columns=fullflexible, mathescape, numbers=none]
Problem:
{problem}

Think step-by-step to reach the solution. Output your reasoning process and final answer using this format:
<think>
[Write your complete reasoning process here, showing each step of your thinking]
</think>

boxed{{answer}}
"""
\end{lstlisting}
\end{tcolorbox}
\begin{tcolorbox}[
    breakable, 
    colback=white!95!gray,
    colframe=gray!50!black,
    rounded corners,
    label={prompt-dot}, 
    title={Prompt for EGSR($s,a$): Generating Trajectory $\tau(s,a)$},
    before upper={\small}
]
\begin{lstlisting}[breaklines=true, xleftmargin=0pt, breakindent=0pt, columns=fullflexible, mathescape, numbers=none]
Problem:
{problem}

Reference Solution:
{solution}

You already know the answer, but it's important to understand the solution in your own way. Like a student who learns by reformulating rather than memorizing, develop your own step-by-step reasoning that leads to a similar conclusion.

IMPORTANT: Do NOT copy the reference solution verbatim. Instead, reconstruct the reasoning path using your own understanding and expression style. Your goal is to demonstrate comprehension by explaining the solution in your natural reasoning pattern.

Use this format:
<think>
[Your original reasoning process here, showing how YOU would reach this solution]
</think>

boxed{{answer}}
\end{lstlisting}
\end{tcolorbox}

\section{Case Study: Emergence of Flawed Reasoning Steps under Answer-Only Guidance in EGSR}
\label{appendix:case_study}
We observe that if EGSR guides self-reformulation using only the final answer (a), the model, despite lower perplexity (PPL) in its outputs, tends to generate flawed reasoning pathways. It appears the model may construct its solution to forcibly arrive at the provided correct answer, a behavior reminiscent of reward hacking.

\begin{tcolorbox}[
    breakable,
    colback=white!95!gray,
    colframe=gray!50!black,
    rounded corners,
    title={Case Study},
    before upper={\small}
]
    \textbf{Question}:
    \par
     Find the maximum value of $\frac{x + 2y + 3}{\sqrt{x^2 + y^2 + 1}}$ over all real numbers $x$ and $y.$
    \tcbline

    \textbf{Solution}: 
    \par
     Because we want to find the maximum value of the expression, we can assume that both $x$ and $y$ are positive; if not, then replacing $x$ and $y$ with $|x|$ and $|y|$ would strictly increase the value of the expression. By Cauchy-Schwarz, $(1^2 + 2^2 + 3^2)(x^2 + y^2 + 1) \ge (x + 2y + 3)^2,$ or $14(x^2 + y^2 + 1) \ge (x + 2y + 3)^2.$ 
    Hence,\[\frac{x + 2y + 3}{\sqrt{x^2 + y^2 + 1}} \le \sqrt{14}.\]Equality occurs when $x = \frac{y}{2} = \frac{1}{3},$ so the maximum value is $\boxed{\sqrt{14}}.$
    \tcbline
    \textbf{Answer}: 
    $\boxed{\sqrt{14}}.$
    \par
    \tcbline 
    
    \textbf{Flawed $\tau(a)$ Example (Correct Answer)}: 
    \par
    [...] (Attempted Lagrange Multipliers, noted system was complex to solve) [...] 
    \textcolor{red}{Since the problem asks for the maximum value..., we can use the fact that the maximum value of a function of the form \(\frac{ax + by + c}{\sqrt{x^2 + y^2 + d}}\) is given by \(\sqrt{a^2 + b^2 + c^2}\).} 
    \textcolor{red!80!black}{\scriptsize (Error: Cited formula is generally incorrect, ignores 'd'.)} 
    Applying this to our function: $a = 1, b = 2, c = 3, d = 1$. The maximum value is: $\sqrt{1^2 + 2^2 + 3^2} = \sqrt{1 + 4 + 9} = \sqrt{14}$. </think>
    Thus, the maximum value is \textcolor{green!50!black}{\(\boxed{\sqrt{14}}\)}. 
    \textcolor{green!50!black}{\scriptsize (Correct answer via incorrect method.)}
    \tcbline

    \textbf{Correct $\tau(s,a)$ Example}: 
    \par
    <think> Applying the Cauchy-Schwarz inequality with vectors \((1, 2, 3)\) and \((x, y, 1)\), we get:
    $(1^2 + 2^2 + 3^2)(x^2 + y^2 + 1) \geq (x \cdot 1 + 2y \cdot 1 + 3 \cdot 1)^2$, This simplifies to: $14(x^2 + y^2 + 1) \geq (x + 2y + 3)^2$. [...]
    Taking the square root of both sides and rearranging yields:
    \[ \sqrt{14} \geq \frac{|x + 2y + 3|}{\sqrt{x^2 + y^2 + 1}} \]
    Since we seek the maximum value, the expression is bounded above by \(\sqrt{14}\).
    Equality holds if and only if the vectors \((1, 2, 3)\) and \((x, y, 1)\) are proportional, i.e., when \(\frac{x}{1} = \frac{y}{2} = \frac{1}{3}\).
    Thus, the maximum value is indeed achievable. </think>
    
    The final answer is: \textcolor{green!50!black}{\(\boxed{\sqrt{14}}\)}. 
\end{tcolorbox}
\fi
\end{document}